\documentclass[lettersize,journal]{IEEEtran}

\usepackage{amsmath}
\usepackage{amsfonts}
\usepackage{booktabs} 
\usepackage[table]{xcolor}
\usepackage{algorithm}
\usepackage{algorithmic}
\usepackage{adjustbox}
\usepackage{tabularx}   
\usepackage[font=footnotesize, labelsep=period]{caption}

\usepackage{amsmath,amsfonts}
\usepackage{algorithmic}
\usepackage{algorithm}
\usepackage{array}
\usepackage[caption=false,font=normalsize,labelfont=sf,textfont=sf]{subfig}
\usepackage{textcomp}
\usepackage{stfloats}
\usepackage{url}
\usepackage{verbatim}
\usepackage{graphicx}
\usepackage{cite}
\begin{document}

\title{InstanceBEV: Unifying Instance and BEV Representation for 3D Panoptic Segmentation}

\author{Feng Li, Zhaoyue Wang, Enyuan Zhang, Mohammad Masum Billah, Yunduan Cui, Kun Xu
}



\maketitle

\begin{abstract}
BEV-based 3D perception has emerged as a focal point of research in end-to-end autonomous driving. However, existing BEV approaches encounter significant challenges due to the large feature space, complicating efficient modeling and hindering effective integration of global attention mechanisms.
We propose a novel modeling strategy, called InstanceBEV, that synergistically combines the strengths of both map-centric approaches and object-centric approaches. Our method effectively extracts instance-level features within the BEV features, facilitating the implementation of global attention modeling in a highly compressed feature space, thereby addressing the efficiency challenges inherent in map-centric global modeling. Furthermore, our approach enables effective multi-task learning without introducing additional module.
We validate the efficiency and accuracy of the proposed model through predicting occupancy, achieving 3D occupancy panoptic segmentation by combining instance information. Experimental results on the OCC3D-nuScenes dataset demonstrate that InstanceBEV, utilizing only 8 frames, achieves a RayPQ of 15.3 and a RayIoU of 38.2. This surpasses SparseOcc’s RayPQ by 9.3\% and RayIoU by 10.7\%, showcasing the effectiveness of multi-task synergy.
\end{abstract}


\section{Introduction}
The design of 3D perception approaches for the environment significantly influences downstream perception-based applications, such as autonomous driving tasks. Current 3D environmental perception methods can be broadly categorized into two main design paradigms: object-centric approaches derived from DETR3D \cite{detr3d}, and map-centric modeling methods like BEVFormer\cite{avidan_bevformer_2022}.
The object-centric paradigm represents an object of interest within a spatial region by a single feature vector, which compresses both the object’s visual characteristics and spatial location. In contrast, the map-centric approach partitions the space into uniform, regular regions, with each region described by a corresponding map feature vector.

Currently, end-to-end paradigms in autonomous driving also follow these two perception approaches. For example, DriveTransformer \cite{jia_drivetransformer_2025} incorporates the object-centric paradigm into a unified Transformer-based \cite{vaswani_attention_2023} end-to-end architecture, while UniAD \cite{uniad} adopts a BEV (Bird’s Eye View) map-centric modeling strategy.
The object-centric design of DETR3D lacks a unified environmental feature representation. Under multi-task settings, it requires assigning separate sets of queries for different tasks \cite{jia_drivetransformer_2025, liu_petrv2_2022}.
In contrast, map-centric methods learn a shared representation on a unified BEV feature map, from which multiple tasks are decoded through task-specific heads. These tasks may include BEV semantic segmentation, object detection, occupancy grid estimation, motion prediction, and more \cite{liu_bevfusion_2023, wang_panoocc_2024, xu_cobevt_2022}. This unified representation framework enhances interpretability.

However, map-centric approaches typically require dense and large-scale BEV representations to describe the overall environment. Current BEV encoders often rely on local-to-global modeling techniques such as CNNs or Deformable Transformers \cite{avidan_bevformer_2022, huang_bevdet_2022, li_fb-bev_2023}.
Occupancy grids, on the other hand, divide the 3D space into uniform cubic volumes, preserving height information more explicitly compared to BEV maps. However, modeling occupancy grids imposes a significantly higher computational burden, which has led to the adoption of spatial sparsification strategies to reduce resource consumption \cite{liu_fully_2024, tang_sparseocc_2024}.
As BEV methods generate a novel top-down perspective from multi-view camera inputs, they inherently involve a level of 3D spatial reasoning. As such, they are also commonly used as intermediate representations for occupancy prediction tasks.
Despite various engineering optimizations, the higher complexity of map-centric methods compared to object-centric ones makes them difficult to scale up, limiting their applicability in real-time downstream tasks such as end-to-end navigation \cite{uniad, jia_drivetransformer_2025, chen_persformer_2022}.

\begin{figure*}[htbp]
  \centering
  \includegraphics[width=\textwidth]{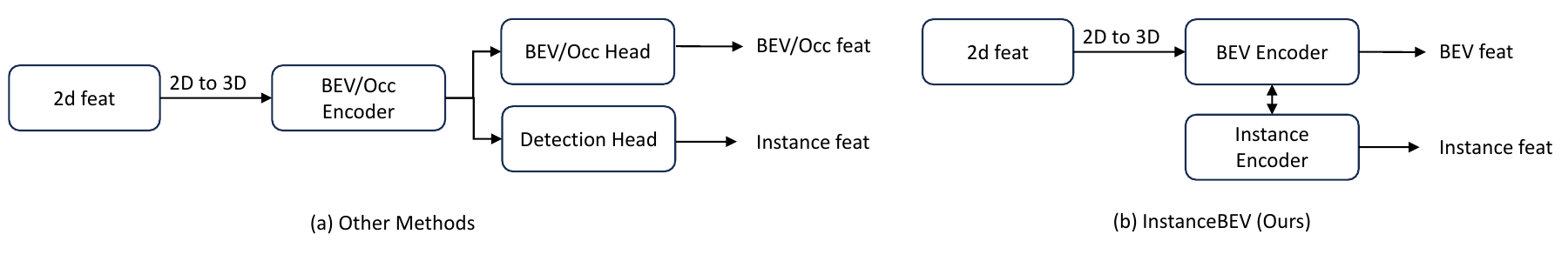}
  \caption{We propose interacting instance features with BEV features during modeling, rather than using a pipeline-like sequential approach. By utilizing instance-level dimensionality reduction, this method enables efficient global modeling, reducing computational complexity while maintaining dense feature representations.}
  \label{fig:framework_contrast}
\end{figure*}

In current autonomous driving systems, perception accuracy is typically measured at the meter level, and object-centric modeling has been widely adopted in 3D perception.This highlights the potential for global attention modeling within a highly compressed latent feature space. Inspired by this insight, we propose a joint modeling approach for BEV features and a compact latent space, enabling the model to maintain fine-grained local resolution while achieving global context modeling in the compressed space. We explicitly define this latent space as a set of \textit{instance queries}, ensuring compatibility with existing perception frameworks. Based on this design, we propose a novel method, InstanceBEV, which successfully unifies the modeling of map-level and instance-level feature spaces.

The concept of jointly modeling different feature spaces has been explored in various contexts, such as multimodal learning, where models combine visual and language representations to enable few-shot or even zero-shot capabilities \cite{li_align_2021, li_blip_2022, radford_learning_2021}.
Extending this idea, SlowFast Networks adopt a similar philosophy by feeding video at different frame rates into two separate pathways: one at a low frame rate to capture spatial semantics and the other at a high frame rate to capture motion with fine temporal resolution \cite{feichtenhofer_slowfast_2019}.

Rather than fusing different modalities or frame rates, InstanceBEV advances this approach by extracting complementary feature representations from the same input data within a unified modality, with each representation encapsulating distinct semantic meanings of the environment.
As illustrated in Figure~\ref{fig:framework_contrast}, our method diverges from existing map-centric models by directly generating a richer set of feature types. This enables the model to provide strong priors for instance-level vision tasks in a multi-task setting.
We validate InstanceBEV through extensive experiments on both occupancy prediction and object detection, demonstrating that our model significantly enhances the understanding of 3D perception in multi-task learning scenarios.

We summarize our main contributions as follows:
\begin{itemize}
\item We propose a modeling approach that extracts complementary feature representations from the same unimodal input, unifying map-level and instance-level visual features. This enables global attention modeling within the BEV space by leveraging the complementarity between different feature spaces.
\item We introduce a bidirectional cross-attention mechanism to efficiently encode single-modal information across distinct feature spaces. In addition, we propose a Residual Prediction strategy for the occupancy head, which significantly improves performance without any cost.
\item Our model supports panoptic segmentation and demonstrates that adding multi-task learning improves occupancy segmentation performance. While enabling direct global attention modeling, our method maintains high runtime efficiency and a compact model size, offering a new perspective for efficient scene representation.
\end{itemize}

\section{Related Work}
\subsection{Object-Centric Perception}
End-to-end object detection, exemplified by DETR \cite{detr}, has become a major research focus within object-centric perception. Deformable DETR \cite{zhu_deformable_2021} introduces deformable attention to accelerate convergence, improve small object detection, and reduce computational complexity.
DAB-DETR \cite{liu_dab-detr_2022} proposes learnable dynamic anchor boxes as object queries, enabling joint optimization of spatial positions and semantic content, which leads to faster convergence and improved detection accuracy.
DN-DETR \cite{li_dn-detr_nodate} enhances training by injecting noisy object and background queries, simulating more complex matching scenarios. This design improves both convergence speed and model stability and performance.

Object-centric 3D perception methods have evolved from DETR3D \cite{detr3d}, which utilizes learnable queries and cross-view attention to sample relevant information from multi-view images, enabling end-to-end 3D bounding box prediction. DD3D \cite{dd3d} introduces a pretraining scheme based on depth estimation that significantly improves the accuracy of 3D perception. Far3D \cite{far3d} leverages high-quality 2D object priors and proposes a perspective-aware aggregation module to enhance the perception of distant objects. PETR \cite{avidan_petr_2022} injects 3D positional encoding directly into 2D image features, effectively avoiding explicit 2D-to-3D transformations. StreamPETR \cite{wang_exploring_2023} proposes a temporally object-centric modeling strategy that successfully addresses the challenge of long-term sequence modeling for object queries. Sparse4D \cite{lin_sparse4d_2023} extends the idea of noisy queries from 2D end-to-end detection into 3D detection and is combined with decoupled attention, achieving impressive performance \cite{li_dn-detr_nodate, zhang_dino_2022}.
VAD \cite{jiang_vad_2023} introduces an instance-level approach to vectorized map construction, making vectorized paradigm for end-to-end autonomous driving possible.

\begin{figure*}[htbp]
  \centering
  \includegraphics[width=\textwidth]{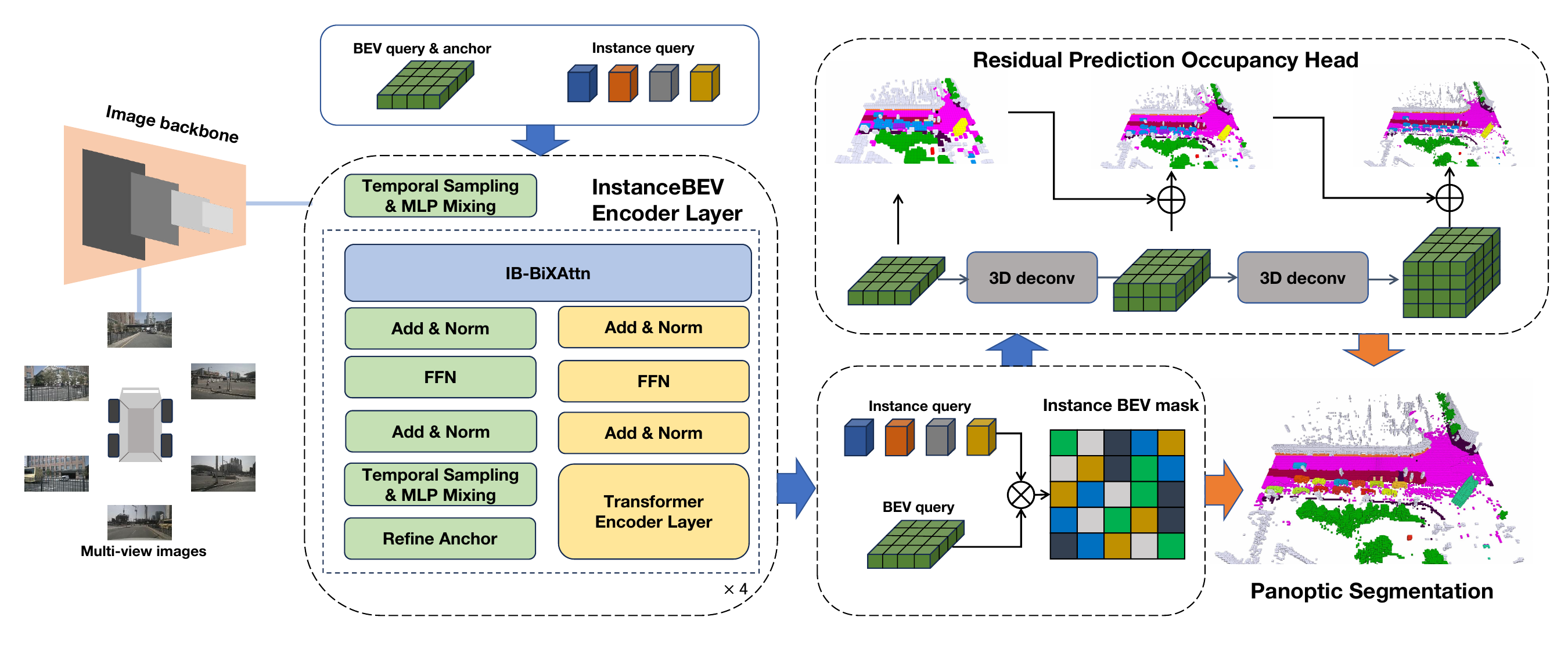}
  \caption{The overall architecture of the proposed InstanceBEV. Images are processed by the backbone and neck to extract multi-level features. The InstanceBEV Encoder fuses BEV and instance features via IB-BiXAttn, where BEV features are transformed from 2D to 3D through sampling and mixing, and instance features are globally modeled by a transformer encoder. The fused features are then decoded to produce instance BEV masks and occupancy segmentation, which are unified into occupancy panoptic segmentation.}
  \label{fig:overall_architecture}
\end{figure*}

\subsection{Map-Centric Perception}
Map-centric perception methods have become a research hotspot, leading to two major representation paradigms: Bird’s Eye View (BEV) maps and occupancy maps. LSS \cite{philion_lift_2020} projects pixel features from each camera into a unified 3D space by casting rays based on camera parameters and discretizing them at multiple depths. These are then aggregated using a PointNet-style operation \cite{qi_pointnet_2017} into BEV pillars to form a BEV feature map. However, the process of lifting pixel-space features into BEV space introduces significant latency, and the lack of accurate depth priors becomes a performance bottleneck.
BEVFusion \cite{liu_bevfusion_2023} accelerates the process by optimizing grid association and feature aggregation in BEV pooling, resulting in a $40\times$ speedup. BEVDepth \cite{li_bevdepth_2023} further improves this by assigning a GPU thread to each frustum feature and replacing the original BEV pooling module with a highly parallelized version, resulting in a speedup of up to $80\times$. It also explicitly introduces a depth estimation module with supervision, leading to perception more accurate.
In contrast, BEVFormer \cite{avidan_bevformer_2022} adopts a backward mapping strategy, where 3D points are projected back into 2D feature maps using camera intrinsics and extrinsics to extract BEV features. FB-BEV \cite{li_fb-bev_2023} combines both forward and backward projection schemes to obtain improved BEV representations.

Compared to BEV maps, occupancy maps preserve height information, thereby directly avoiding height-related information loss at the modeling level. With dataset support provided by OCC3D \cite{tian_occ3d_2023} and OpenOcc \cite{tong_scene_2023}, a growing number of occupancy-based modeling methods have emerged \cite{wang_occsora_2024, yu_flashocc_2023, tang_sparseocc_2024, wang_opus_2024}.
PanoOcc \cite{wang_panoocc_2024} employs a coarse-to-fine strategy combined with an integrated occupancy sparsification module to achieve efficient occupancy modeling. TPVFormer \cite{huang_tri-perspective_2023} reformulates the 3D modeling problem into a tri-perspective view problem, significantly reducing the modeling complexity by orders of magnitude.
SparseOcc \cite{liu_fully_2024} proposes the first fully sparse occupancy modeling method. To address the common issue of overly thick surface predictions that artificially inflate IoU metrics, it introduces a novel lidar ray-based evaluation metric called RayIoU, which considers only the first visible voxel along the ray for evaluation.

\section{Method}
This section begins with an overview of the proposed model, followed by the introduction of a bidirectional cross-attention module that jointly encodes BEV and instance feature spaces. Next, we detail the temporal modeling framework within the BEV domain. Finally, we present the Residual Prediction Occupancy Head along with the associated instance query supervision strategy.

\subsection{Overall Architecture}
Our overall architecture is illustrated in Figure~\ref{fig:overall_architecture}. The proposed modeling method consists of two branches: BEV features to ensure the fidelity of local information, and Instance features to achieve high compression performance of the model.
In the encoder, BEV queries are first initialized from image features using temporal sampling and an MLP mixing. Subsequently, the BEV queries and instance queries are jointly modeled and fused through multiple layers of the InstanceBEV encoder.
The output instance queries and BEV queries are matched via cosine similarity to obtain instance masks in the BEV space. Both the BEV queries and the matched instance queries are then input into a Residual Prediction Occupancy Head to produce occupancy predictions.
Finally, the model utilizes the instance masks to perform panoptic segmentation in the 3D occupancy space.

\subsection{Instance-BEV Multi-Head Bidirectional Cross Attention}
The attention mechanism is recognized as an effective method for global modeling. The concept of bidirectional cross-attention has been adopted in multi-modal settings~\cite{lu_vilbert_2019,hiller_perceiving_2024}.
Traditional bidirectional cross-attention typically requires two separate cross-attention operations, resulting in the computation of two attention matrices. To enhance efficiency in unimodal scenarios, we propose a compact variant that just computes a single attention matrix while simultaneously updating both the instance queries and BEV queries.
Our objective is for the instance queries to learn instance-level representations within the BEV space, while the BEV queries gain the capability to differentiate instance semantics at each spatial location.

\textbf{Instance-BEV Multi-Head Bidirectional Cross Attention (IB-BiXAttn).}  
As illustrated in Fig.~\ref{fig:overall_architecture}, we adopt a bidirectional cross-attention mechanism to enable feature transformation between Instance features and BEV features. This facilitates instance-level dimensionality reduction in the BEV space.
Let the instance features be denoted as $Q^I \in \mathbb{R}^{n_i \times c}$ and the BEV features as $Q^B \in \mathbb{R}^{n_b \times c}$, where $n_i$ and $n_b$ are the numbers of instance and BEV queries, and $c$ is the channel dimension.
The linearly projected queries ${Q}^I$ and ${Q}^B$ are first split evenly along the channel dimension into $N$ groups:
\[
[\mathcal{Q}^I_0, \mathcal{Q}^I_1, ..., \mathcal{Q}^I_{N-1}],\quad [\mathcal{Q}^B_0, \mathcal{Q}^B_1, ..., \mathcal{Q}^B_{N-1}].
\]
For each head $i$, we compute the shared attention score matrix as:
\begin{align}
    \text{Attention Score}_i &= \frac{\mathcal{Q}^I_i \cdot (\mathcal{Q}^B_i)^T}{\sqrt{c/N}}.
\end{align}

Then, both the instance queries and BEV queries are updated using the same attention scores:
\begin{align}
    Q^I &= \text{Concat}(head^I_0, head^I_1, ..., head^I_{N-1}), \\
    Q^B &= \text{Concat}(head^B_0, head^B_1, ..., head^B_{N-1}),
\end{align}
where
\begin{align}
head^I_i &= \text{Linear}(\text{softmax}(\text{Attention Score}_i) \cdot \mathcal{Q}^B_i), \\
head^B_i &= \text{Linear}(\text{softmax}(\text{Attention Score}_i^T) \cdot \mathcal{Q}^I_i).
\end{align}

Here, $\text{Attention Score}_i^T$ denotes the transpose of the score matrix across the last two dimensions. When the instance queries are replaced by BEV queries, IB-BiXAttn degenerates into standard self-attention over BEV queries.

\textbf{Instance queries significantly reduce attention complexity for BEV.}  
Traditional BEV modeling typically relies on local computation mechanisms such as CNNs and Deformable Attention, whose computational complexity scales linearly with the number of queries. This design choice avoids the high cost of global modeling over the entire BEV space. 
In contrast, our IB-BiXAttn reduces the BEV queries to a smaller set of instance queries, enabling direct global modeling via attention mechanisms. Computing global self-attention over BEV queries requires a complexity of $\mathcal{O}(n_b \times n_b)$. By introducing instance queries, we instead compute a cross-attention score matrix and apply global attention in the reduced instance space, resulting in a total complexity of $\mathcal{O}(n_b \times n_i) + \mathcal{O}(n_i \times n_i)$.
When $n_i \ll n_b$, this leads to a significant reduction:  
$
\mathcal{O}(n_b \times n_i) + \mathcal{O}(n_i \times n_i) \ll \mathcal{O}(n_b \times n_b).
$

\subsection{Temporal Module}
\textbf{Temporal Sampling.}  
We construct our temporal module from multi-view sequential images. Suppose there are $N$ surround-view cameras and the model takes $T$ historical timestamps as input. The model input thus consists of $T \times N$ images:  
\begin{equation}
    \{(I_1^t, I_2^t, ..., I_N^t),\ t \in \{0, 1, ..., T-1\} \}.
\end{equation}
For each BEV anchor, we generate spatial sampling points in the current vehicle coordinate system. Specifically, the $i$-th BEV anchor produces $T$ sets of sampling points:  
\begin{equation}
\{P_i^t,\ t \in \{0,1,...,T-1\} \}.
\end{equation}
Then,  we transform these sampling points from the current vehicle coordinate frame to each historical ego frame to obtain the temporal sample points using the current ego pose $E^0$ and history ego pose at prev $t$ frames $E^{t}$:  
\begin{equation}
P_i = \{(E^t)^{-1} E^0 P_i^t,\ t \in \{0,1,...,T-1\} \}.
\end{equation}
These transformed points are subsequently projected into the image planes using the extrinsic calibration between vehicle and camera and the camera intrinsics. Bilinear interpolation is used to extract the corresponding image features.

\textbf{MLP Mixing.}  
Suppose a BEV pillar has a BEV feature $Q^B_i \in \mathbb{R}^{1 \times c}$ and generates $n$ sampling points, where the sampled image features have a channel dimension of $c_p$. Then the sampling features for this BEV pillar are denoted as $S_i \in \mathbb{R}^{n \times c_p}$.  
To aggregate multiple sampling point features into a unified BEV feature, we adopt the MLP-Mixer~\cite{tolstikhin_mlp-mixer_2021} for each BEV pillar. Specifically, we perform point mixing and channel mixing across the point and feature dimensions, respectively. The output is then flattened and transformed to match the BEV query dimensionality via a linear layer. Formally, the process is defined as:
\begin{align}
    S_i^{pm} &= \text{ReLU}(\text{DNorm}(W_{pm} S_i + b_{pm})), \\
    S_i^{cm} &= \text{ReLU}(\text{DNorm}(S_i^{pm} W_{cm} + b_{cm})), \\
    Q^B_i &= \text{Linear}(\text{Flatten}(S_i^{cm})) + Q^B_i.
\end{align}
Here, $\text{DNorm}$ denotes a dual-dimensional normalization that applies Layer Normalization over both the point and channel dimensions.

\textbf{Refine Anchor.}  
Since we adopt a dense BEV pillar representation, we do not adjust the pillar width or length. Instead, we refine only the height in the Refine module. Specifically, we update the pillar heights $H \in \mathbb{R}^{n_b \times 1}$ using the BEV features $Q^B \in \mathbb{R}^{n_b \times c}$ obtained from the sampling and mixing modules.  
The height is defined as a normalized value between 0 and 1, representing the relative position between the predefined minimum and maximum heights of the perception range. The height refinement is computed as:
\begin{equation}
    H_{\text{new}} = \text{Sigmoid}(\text{Linear}(Q^B)).
\end{equation}

\subsection{Residual Prediction Occupancy Head}
Residual Prediction refers to a mechanism where the current layer is designed to estimate the residual error of the previous layer. Residual Connection proposed by ResNet~\cite{he_deep_2015} is widely use in most of Deep Neural Networks. However, Dense Connection proposed by DenseNet~\cite{huang_densely_2018} is proved a more effective way to reuse the features while suffering large training memory consumption. The Residual Predition is a stratage which make Residual Connection Decoder performs as well as Dense Connection while without more training memory. We use the ResNet and DenseNet as the example to prove the ResNet with Residual Prediction equals to the DenseNet.  

For DenseNet, all previous layer outputs are concatenated. Suppose the concatenated feature is  $X = \text{Concat}(x_n, x_{n-1}, ..., x_0),$ where $x_n$ represents the output feature of the $n$-th layer. A linear layer is then applied to compute the classification output, which is equivalent to applying a linear transformation to each layer's output individually and summing the results. Mathematically, this can be expressed as:  
\begin{equation}
\begin{aligned} 
    \text{Linear}(X) &= A X + b \\
    &= A_n x_n + A_{n-1} x_{n-1} + \cdots + A_0 x_0 + b \\
    &= \text{Linear}_n(x_n) + \cdots + \text{Linear}_0(x_0).
\end{aligned}  
\end{equation}

Here, $X$ denotes the concatenated vector output from DenseNet, constructed from the outputs of previous $n$ layers $x_0, x_1, ..., x_n$. $A$ and $b$ are the weights and bias of the linear layer, respectively. This equation follows from the block-wise computation property of matrix multiplication.  
This equivalence implies that the final prediction of DenseNet can be viewed as the sum of predictions from all individual layers. Therefore, if residual blocks and dense blocks are equivalent in terms of input features, we can construct DenseNet directly by employing Residual Prediction on ResNet.

In the $(n+1)$-th layer of ResNet, the input to the residual block can be formulated as:
\begin{equation}
\begin{aligned} 
    x_n &= F_{n-1}(x_{n-1}) + x_{n-1} \\
    &= F_{n-1}(x_{n-1}) + F_{n-2}(x_{n-2}) + \cdots,
\end{aligned}  
\end{equation}
where $F_i(\cdot)$ denotes the nonlinear transformation in the $i$-th residual block. This recursive formulation shows that $x_n$ aggregates features from all previous layers, similar to DenseNet.

In DenseNet, the transition layer is used to reduce the number of channels to prevent feature and parameter explosion as the depth increases. We assume that the transition layer computation is channel-wise separable, the computation at the $n$-th layer can be rewritten as:
\begin{equation}
\begin{aligned} 
    x_n &= \text{Transition Layer}(\text{Concat}(F_{n-1}(x_{n-1}), x_{n-1}, ..., x_0)) \\
    &= T_{n-1}[F_{n-1}(x_n-1),x_{n-1}] + T_{n-2}[x_{n-2}] + \cdots
\end{aligned}  
\end{equation}
Here, $T_i(\cdot)$ denotes a learnable transformation applied to the $i$-th layer's output implemented as channel-wise operations.

It is clear that, with appropriate design of the $T_i$ functions, the above formulation can be made equivalent to the residual connection together with Residual Prediction.

\subsection{Instance BEV Mask}
Inspired by Mask2Former~\cite{cheng_masked-attention_2022}, instance queries are employed to predict instance masks in the BEV for instance supervision. Cosine similarity is computed between instance queries and BEV queries to generate soft association scores. Each BEV feature vector is then assigned to the instance query with the highest similarity, producing an instance-aware BEV map. This representation, combined with the encoder’s BEV features, is subsequently fed into the Occupancy Decoder for final occupancy prediction.

\subsection{Loss Function}
For the occupancy task, we employ a loss function that combines the weighted cross-entropy loss for classification, the Dice loss \cite{milletari_v-net_2016}, and the Lovasz loss \cite{berman_lovasz-softmax_2018}: 
\begin{equation}
\mathcal{L}_{\text{occ}}= \lambda_1 \mathcal{L}_{\text{ce}}+\lambda_2 \mathcal{L}_{\text{dice}}+\lambda_3 \mathcal{L}_{\text{Lovasz}} 
\end{equation}
Here, we set $\lambda_1=\lambda_3=1$ and $\lambda_2=0.3$. 

For Instance BEV mask task, we use the Mask2Former loss, which uses the focal loss \cite{lin_focal_2020} for instance classification, the binary cross-entropy loss and the Dice loss for Instance BEV binary mask prediction:
\begin{equation}
\mathcal{L}_{\text{det}}= \lambda_4\mathcal{L}_{\text{focal}}+\lambda_5 \mathcal{L}_{\text{bce}}+\lambda_6 \mathcal{L}_{\text{dice}} 
\end{equation}
Here, we set $\lambda_4=\lambda_5=\lambda_6 = 1$. 

Thus, the total training loss for InstanceBEV is $\mathcal{L}= \mathcal{L}_{\text{occ}} + \mathcal{L}_{\text{det}}$.

\section{Experiments}
\begin{table*}[h]
\centering
\small
\begin{tabular}{lcccc>{\columncolor{gray!20}}c>{\columncolor{gray!20}}cc}
\toprule
\textbf{Method} & \textbf{Backbone} & \textbf{Input Size} & \textbf{Epoch} & \textbf{mIoU} & \textbf{RayIoU$_{1m,2m,4m}$} & \textbf{RayIoU} & \textbf{FPS} \\
\midrule
BEVFormer (4f) & R101 & 1600×900 & 24 & 39.2 & 26.1 / 32.9 / 38.0 & 32.4 & 3.0 \\
RenderOcc      & Swin-B & 1408×512 & 12 & 24.4 & 13.4 / 19.6 / 25.5 & 19.5 & - \\
SimpleOcc      & R101 & 672×336 & 12 & 31.8 & 17.0 / 22.7 / 27.9 & 22.5 & 9.7 \\
BEVDet-Occ (2f) & R50 & 704×256 & 90 & 36.1 & 23.6 / 30.0 / 35.1 & 29.6 & 2.6 \\
BEVDet-Occ-Long (8f) & R50 & 704×384 & 90 & \textbf{39.3} & 26.6 / 33.1 / 38.2 & 32.6 & 0.8 \\
FB-Occ (16f)       & R50 & 704×256 & 90 & 39.1 & 26.7 / 34.1 / 39.7 & 33.5 & 10.3 \\
SparseOcc (8f)  & R50 & 704×256 & 24 & 30.1 & 28.0 / 34.7 / 39.4 & 34.0 & \textbf{17.3} \\
SparseOcc (16f)   & R50 & 704×256 & 24 & 30.6 & 29.1 / 35.8 / 40.3 & 35.1 & 12.5 \\
\midrule
InstanceBEV (8f)  & R50 & 704×256 & 24 & 30.1 & 32.8 / 39.0 / 42.7 & 38.2 &  11.1 \\
InstanceBEV (16f)  & R50 & 704×256 & 24 & 30.4 & \textbf{33.2 / 39.2 / 43.0} & \textbf{38.5} & 7.7\\
\bottomrule
\end{tabular}
\caption{\textbf{Occupancy prediction performance on Occ3D-nuScenes.} Baseline results are from SparseOcc report \cite{liu_fully_2024}.}
\label{tab:occupancy_comparison}
\end{table*}

\begin{table*}[htbp]
\centering
\small
\setlength{\tabcolsep}{2pt}
\renewcommand{\arraystretch}{1.1}
\begin{tabular}{l|cc|
>{\centering\arraybackslash}p{1.5em} >{\centering\arraybackslash}p{1.5em}
>{\centering\arraybackslash}p{1.5em} >{\centering\arraybackslash}p{1.5em}
>{\centering\arraybackslash}p{1.5em} >{\centering\arraybackslash}p{1.5em}
>{\centering\arraybackslash}p{1.5em} >{\centering\arraybackslash}p{1.5em}
>{\centering\arraybackslash}p{1.5em} >{\centering\arraybackslash}p{1.5em}
>{\centering\arraybackslash}p{1.5em} >{\centering\arraybackslash}p{1.5em}
>{\centering\arraybackslash}p{1.5em}
>{\centering\arraybackslash}p{1.5em}
>{\centering\arraybackslash}p{1.5em}
>{\centering\arraybackslash}p{1.5em}
>{\centering\arraybackslash}p{1.5em} >{\centering\arraybackslash}p{1.5em}
}
\toprule
& & & \multicolumn{17}{c}{Per-class RayIoU} \\
Method & \rotatebox{90}{mIoU} & \rotatebox{90}{RayIoU} &
\rotatebox{90}{others} & \rotatebox{90}{barrier} & \rotatebox{90}{bicycle} & \rotatebox{90}{bus} & \rotatebox{90}{car} & \rotatebox{90}{cons.veh.} &
\rotatebox{90}{motor.} & \rotatebox{90}{pedes.} & \rotatebox{90}{tfc.cone} & \rotatebox{90}{trailer} & \rotatebox{90}{truck} &
\rotatebox{90}{drv.surf.} & \rotatebox{90}{other flat} & \rotatebox{90}{sidewalk} & \rotatebox{90}{terrain} &
\rotatebox{90}{manmade} & \rotatebox{90}{vegetation} \\
\midrule
BEVFormer & 23.7 & 33.7 & 5.0 & 42.2 & 18.2 & 55.2 & 57.1 & 22.7 & 21.3 & 31.0 & 27.1 & \textbf{30.7} & 49.4 & 58.4 & 30.4 & 29.4 & 31.7 & 36.3 & 26.5 \\
FB-Occ & 27.9 & 35.6 & \textbf{10.5} & 44.8 & 25.6 & 55.6 & 51.7 & 22.6 & 27.2 & \textbf{34.3} & 30.3 & 23.7 & 44.1 & 65.5 & 33.3 & 31.4 & 32.5 & \textbf{39.6} & \textbf{33.3} \\
\midrule
InstanceBEV(8f) & \textbf{30.1} & \textbf{38.2} & 9.8 & \textbf{47.4} & \textbf{29.6} & \textbf{60.5} & \textbf{57.4} & \textbf{24.4} & \textbf{32.0} & 30.6 & \textbf{33.5} & 27.2 & \textbf{50.4} & \textbf{68.7} & \textbf{38.1} & \textbf{34.7} & \textbf{35.4} & 38.8 & 30.0 \\
\bottomrule
\end{tabular}
\caption{\textbf{Per-class RayIoU performance on Occ3D-nuScenes validation dataset.} All models were trained without camera mask.}
\label{tab:rayiou}
\end{table*}

\subsection{Experimental setup}
\textbf{Dataset.} Occ3D-nuScenes~\cite{tian_occ3d_2023} is a large-scale dataset for autonomous driving. It utilizes six surround-view cameras to capture full 360° observations around the vehicle. The dataset contains 700 training scenes and 150 validation scenes. The occupancy volume spans from \(-40\,\text{m}\) to \(40\,\text{m}\) along the X and Y axes, and from \(-1\,\text{m}\) to \(5.4\,\text{m}\) along the Z axis, which is voxelized into a grid of size \(200 \times 200 \times 16\). The semantic occupancy labels are categorized into 17 classes, including 16 foreground semantic classes and one unknown class.

\textbf{Implementation Details.} The original resolution of the input camera images from dataset is $1600 \times 900$. During training, images are randomly scaled within a ratio range of (0.38, 0.55), followed by cropping to a fixed size of 704×256. Each data augmentation strategy is applied with a probability of 50\%, including random brightness adjustment, contrast adjustment, hue shift, and channel permutation. The images are normalized using the standard ImageNet mean and standard deviation before being fed into the network. Following common practice, we employ a ResNet-50~\cite{he_deep_2015} as the backbone and extract multi-scale feature maps from stages conv2\_x, conv3\_x, conv4\_x, and conv5\_x, corresponding to different image features resolution. The features from these stages, with output channels of 256, 512, 1024, and 2048 respectively, are unified to 256 channels using a Feature Pyramid Network~\cite{lin2017feature}, yielding our multi-level image features. The encoder consists of 4 layers, and the BEV representation is set to a resolution of \(100 \times 100\). We adopt two types of positional encodings in IB-BiXAttn: 2D sinusoidal positional encoding for BEV queries, which encodes fixed positional information, and learnable positional encoding for instance queries. The number of instance queries is set to 200 according the query ablation study and the number of BEV queries is $100 \times 100$. We divide the multi-level image features along the channel dimension into 4 groups, resulting in each group having feature channels $c_p = 64$. 3D sample point are projected to image reference points on multi-level image features, and bilinear interpolation is used to calculate these reference points feature. The sampled features $S \in \mathbb{R}^{n_b \times n \times c_p}$ are obtained by querying the historical image features with BEV queries $Q^B \in \mathbb{R}^{n_b \times c}$. The total number of sampled points is $n = 4 \times T \times n_p$, where $T$ is the number of input frames and $n_p$ is a hyperparameter (set to 4 when $T = 8$). Mixing module applies point and channel mixing to the sampled points, followed by flattening and a linear projection: 
\begin{enumerate}
    \item Point mixing, linear module transforming $S$ into $S^{pm} \in \mathbb{R}^{n_b \times n_{pm} \times c_p}$,
    \item Channel mixing, linear module transforming $S^{pm}$ into $S^{cm} \in \mathbb{R}^{n_b \times n_{pm} \times c_{cm}}$.
\end{enumerate}
We set the mixing parameters as $n_{pm} = T \times n_p$ and $c_{cm} = c_p$.
We use the AdamW~\cite{loshchilov2017decoupled} optimizer with an initial learning rate of \(2 \times 10^{-4}\). The learning rate is decayed at epochs 22 and 24 by a multiplicative factor of 0.2. The model is trained for a total of 24 epochs with a batch size of 8. Similar to SparseOcc~\cite{liu_fully_2024}, we do not apply camera mask during training.

\subsection{RayIoU and RayPQ Metric}
RayIoU, first introduced in SparseOcc~\cite{liu_fully_2024}, addresses the issue of mIoU inflation from thick surface predictions by evaluating on ray-based queries instead of voxels; RayPQ builds on this idea by extending the panoptic quality (PQ) metric to the ray level, computing true positives accordingly.

RayIoU is the mean IoU by ray casting. It computes visible IoU by casting rays from historical ego positions through both predicted and ground truth occupancy. RayIoU computes mIoU by considering only the first intersected voxel along each ray, while allowing a certain depth tolerance. Unlike voxel-based mIoU metric, RayIoU is a ray-based metric, capturing view-dependent quality of scene understanding. 

Similar to RayIoU and rooted in the widely adopted PQ metric, RayPQ is formulated as the multiplication of segmentation quality (SQ) and recognition quality (RQ). RayPQ follows the true positive (TP) definition used in RayIoU, where a predicted instance $p$ is matched with a ground-truth instance $g$ if their IoU exceeds 0.5. Based on these matches, RayPQ is computed as the product of SQ and RQ, following the standard formulation of panoptic quality but adapted to ray-based occupancy evaluation.

\subsection{Main results}
\textbf{Quantitative Performance.} Comparison data in table~\ref{tab:occupancy_comparison} and table~\ref{tab:rayiou} are reported by SparseOcc\cite{liu_fully_2024}. It presents a comprehensive comparison between our method and existing baselines on the Occ3D-nuScenes dataset. In table~\ref{tab:occupancy_comparison}, our method achieves a RayIoU of 38.2 just using 7 historical frames, outperforming other approaches, including FB-Occ (16f) which training with camera mask, attains a RayIoU of 33.5 (-4.7), and SparseOcc (16f) which training without camera mask, with a RayIoU of 35.1 (-3.1). In addition to superior accuracy, our method demonstrates enhanced real-time efficiency compared to dense BEV-based methods. In the Table~\ref{tab:rayiou}, InstanceBEV demonstrates a clear advantage over other dense BEV models trained without camera visible mask supervision. In particular, it achieves superior RayIoU performance on road-participating classes such as car, bus, motorcycle, and bicycle. 

\begin{figure*}[htbp]
  \centering
  \includegraphics[width=0.95\textwidth]{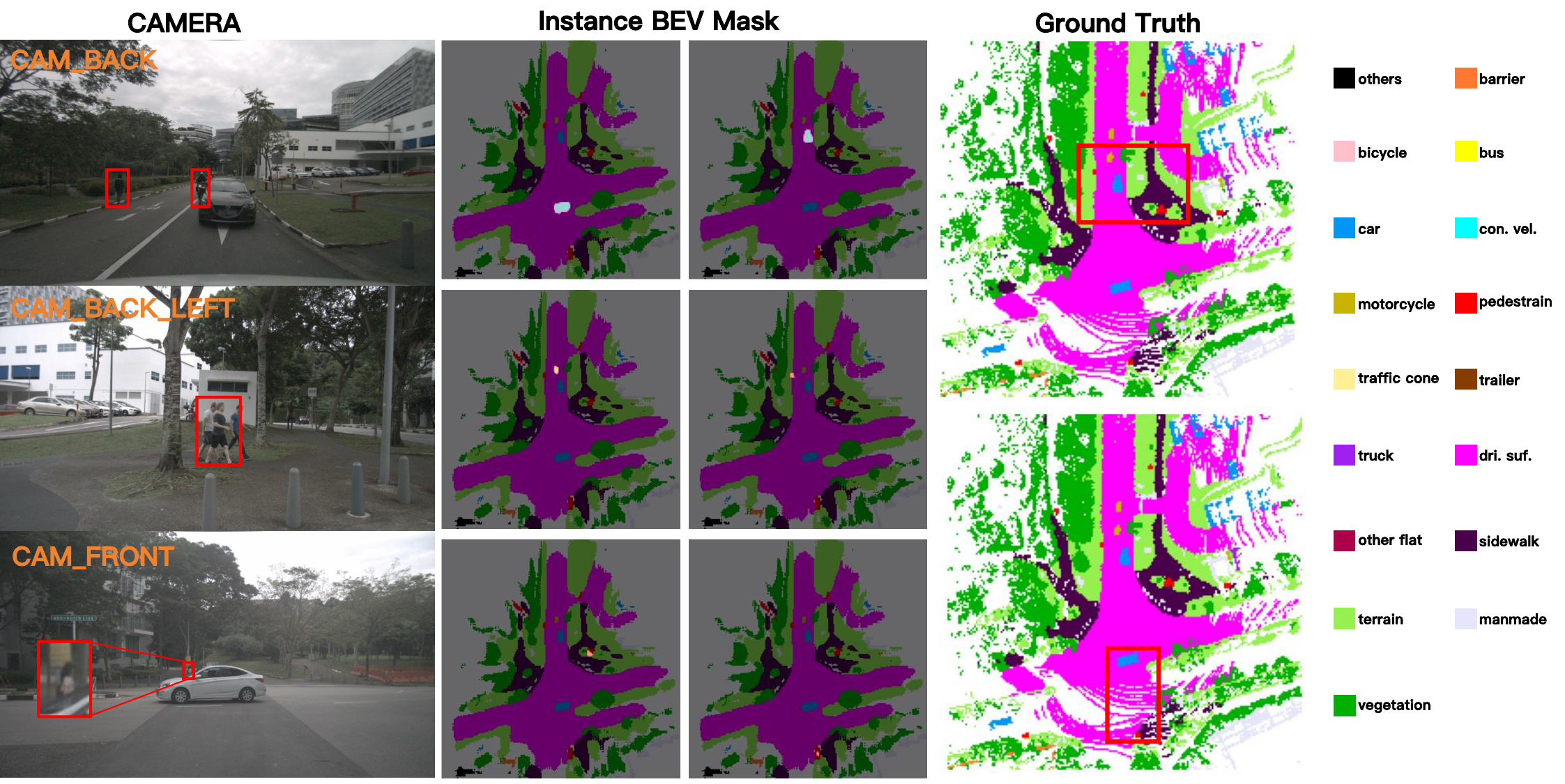}
  \caption{\textbf{Visualization of InstanceBEV.} We visualize the predicted occupancy in the BEV perspective, overlaid with the corresponding Instance BEV masks. The InstanceBEV representation effectively demonstrates multi-task capability by leveraging distinct feature spaces to capture both semantic and instance-level information. The visualization shows the model’s ability to localize and differentiate individual objects within the 3D scene while maintaining accurate semantic understanding of the environment.}
  \label{fig:instance bev mask}
\end{figure*}

\textbf{Visualization.} Our method not only produces occupancy segmentation but also predicts instance-level masks in the BEV space. We visualize a sample output occuapncy segmantation combines with Instance BEV mask on BEV, which could represent InstanceBEV's multi-task ability. As illustrated in Figure~\ref{fig:instance bev mask}, InstanceBEV accurately captures and distinguish different large dynamic objects such as vehicles, as well as smaller instances like pedestrians, successfully distinguishing between instances. By associating the predicted occupancy with instance masks, we can achieve panoptic segmentation in 3D space directly, eliminating the need for an additional instance head. Further qualitative results for occupancy prediction and occupancy panoptic segmentation are available in the next section.

\subsection{Ablation studies}
\textbf{The Efficency of IB-BiXAttn.} The number of instance queries plays a critical role in balancing the attention computation complexity and overall model performance. We conduct ablation studies to investigate the impact of the number of instance queries on both accuracy and inference latency. As shown in Table~\ref{tab: query ablation}, performance begins to saturate as the number of queries approaches 100 and peaks at 200 queries. Notably, our proposed instance-level dimensionality-reduced attention demonstrates sustained real-time performance as the number of queries increases. Even with 200 instance queries, the inference latency of update both instance and BEV queries only increase to 3.28\,ms, which demonstrate the real-time global modeling.

\begin{figure}[tbp]
  \centering
  \includegraphics[width=0.4\textwidth]{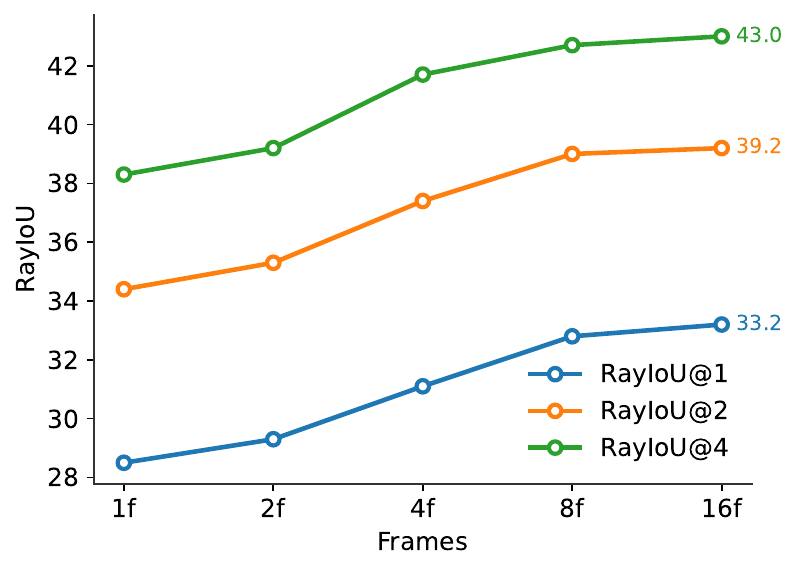}
  \caption{\textbf{Temporal modeling performance.} Performance improves as the number of input frames increases, exhibiting signs of saturation at 8 frames and reaching its peak at 16 frames.}
  \label{fig:temporal}
\end{figure}

\textbf{Temporal Modeling.} 
As shown in Figure~\ref{fig:temporal}, we validate the effectiveness of our temporal modeling strategy. Experimental results indicate that as more temporal frames are incorporated, our sampling and mixing approach effectively fuses temporal features, resulting in consistent performance improvements. With 8 input frames, the model performance reaches saturation, suggesting the sufficiency of temporal context under our fusion strategy.

\begin{table}[tbp]
\centering
\small
\begin{tabular}{ccc}
\toprule
Query Number & RayIoU & IB-BiXAttn Latency \\
\midrule
20 & 33.1 & \textbf{2.44 ms}\\
50 & 33.3 & 2.55 ms\\
100 & 33.6 & 2.78 ms\\
200 & \textbf{33.7} & 3.28 ms\\
\bottomrule
\end{tabular}
\caption{\textbf{Query Ablation Study.} The proposed instance-level dimensionality-reduced attention demonstrates sustained real-time performance as the number of queries increases. Latency is measured on an NVIDIA A100 GPU.}
\label{tab: query ablation}
\end{table}

\begin{table}[tbp]
\centering
\small
\begin{tabular}{ccc}
\toprule
Method & RayIoU & RayPQ \\
\midrule
SparseOcc(8f) \dag & 35.0 & -\\
InstanceBEV (8f) \dag & 37.9 & -\\
\midrule
SparseOcc(8f) & 34.5 (-0.5) & 14.0\\
InstanceBEV (8f) & \textbf{38.2 (+0.3)} & \textbf{15.3}\\
\bottomrule
\end{tabular}
\caption{\textbf{Ablation study for instance supervision.} Results demonstrate that InstanceBEV effectively accommodates multi-task. By leveraging instance queries and BEV queries to predict instance masks in the BEV space, our method achieves panoptic occupancy segmentation. The symbol $\dag$ indicates training without instance supervision.}
\label{tab: pano}
\end{table}

\begin{figure*}[h]
\centering
\includegraphics[width=\textwidth]{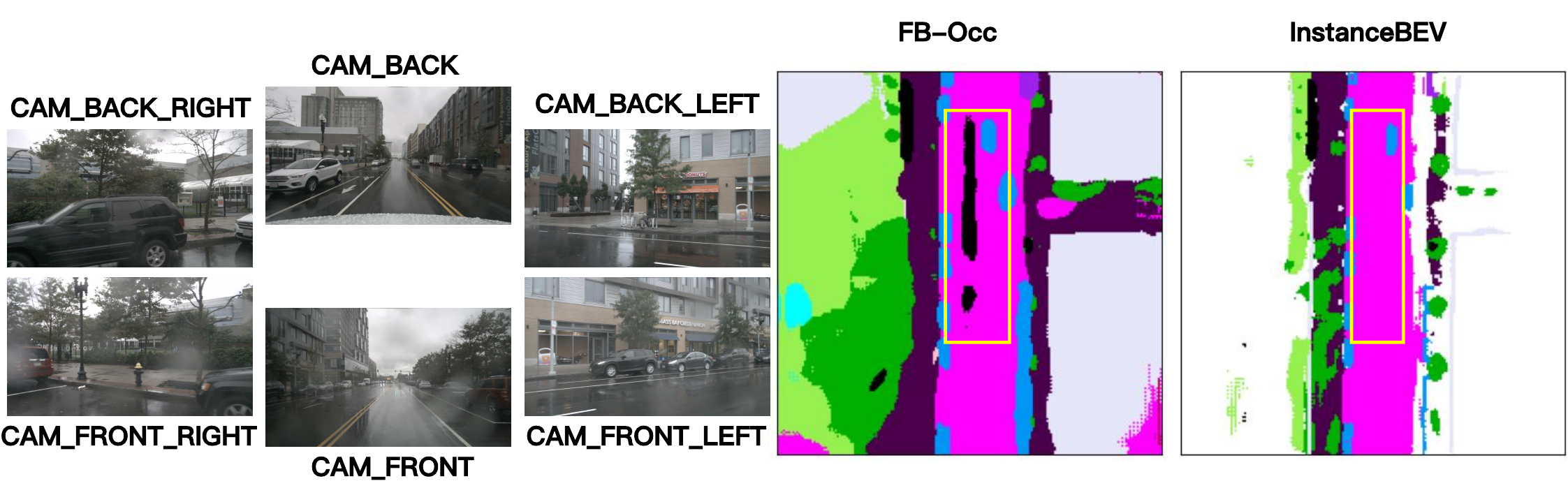}
\caption{\textbf{Visualization BEV View from OCC3D-nuScenes validation set.} FB-Occ which uses camera mask in training period causes plenty of false positive predictions on the drivable surface.}
\label{fig: risk}
\end{figure*}

\textbf{Instance BEV Mask Supervision.}
We conduct an ablation study on the BEV mask supervision from instance queries to evaluate its impact. As shown in Table \ref{tab: pano}, multi-task supervision enhances the model’s understanding of the 3D perceptual space. The comparison data is sourced from SparseOcc\footnote{\url{https://github.com/MCG-NJU/SparseOcc/tree/v1.0}}. Notably, our method performs panoptic segmentation by directly supervising the instance queries within the encoder, eliminating the need for an additional instance decoder.
Our approach achieves a RayPQ of 15.3 and enhances RayIoU to 38.2, surpassing SparseOcc’s RayPQ of 14.0 (an improvement of +1.3, or 9.3\%) and RayIoU of 34.5 (an increase of +3.7, or 10.7\%).

\textbf{The Effectiveness of Residual Prediction.} 
We theoretically demonstrate that using Residual Prediction with a residual connection is equivalent to employing a dense connection. We conduct an ablation study by removing the Residual Prediction branch and directly comparing residual and dense connection. The results in Table~\ref{tab: decoder} indicate that our Residual Prediction strategy enables residual to achieve performance comparable to that of the dense connection, while significantly reducing GPU memory consumption during training. This validates the effectiveness of Residual Prediction in achieving high performance without any cost. Results indicate that Residual Prediction improves the residual connection without addtional parameters and GPU memory.

\textbf{Global vs. Local Feature Modeling.} 
To evaluate the effectiveness of global modeling with attention, we conduct ablation by removing the IB-BiXAttn module from our architecture. This also eliminates the self-attention computations in the instance query branch. As shown in Table~\ref{tab: encoder}, we replace the attention-based encoder with commonly used local modeling methods in BEV frameworks, including CNNs and Deformable Attention. The results demonstrate that our proposed global modeling method outperforms traditional local computation approaches.

\begin{table}[tbp]
\centering
\small
\begin{tabular}{ccc}
\toprule
Occupancy Head & RayIoU & Training Memory \\
\midrule
Residual & 32.5 & \textbf{6.4 G}\\
Dense & \textbf{33.7} & 14.7 G\\
Residual + Residual Prediction & \textbf{33.7} & \textbf{6.4 G}\\
\bottomrule
\end{tabular}
\caption{\textbf{Residual Prediction Ablation.} Results indicate that residual prediction improves the residual connection without addtional parameters and GPU memory. Memory is measured with batch size of one on each NVIDIA A100 GPU.}
\label{tab: decoder}
\end{table}

\begin{table}[tp]
\centering
\small
\begin{tabular}{cc} 
\toprule
Method & RayIoU \\
\midrule
CNN & 32.8   \\
Deformable Attention & 33.2   \\
IB-BiXAttn + Self-Attn & \textbf{33.7} \\
\bottomrule
\end{tabular}
\caption{\textbf{Comparison with other BEV computation modules.} We compare our proposed global modeling strategy, IB-BiXAttn combined with Self-Attention, against mainstream BEV modeling approaches that primarily rely on local computation and sparse computation. Our method outperforms other methods, demonstrating the effectiveness of global context aggregation in BEV space.}
\label{tab: encoder}
\end{table}

\begin{table}[tp]
\centering
\small
\begin{tabular}{lcc}
\toprule
Method & Resolution & Memory \\
\midrule
SparseOcc(8f) & 32000 & 6.8 G\\
BEVDet-Occ-Long (8f) & $200 \times 200 \times 16$ & 6.3 G \\
FB-Occ(16f) & $100 \times 100 \times 8$ & 5.1 G\\
InstanceBEV(8f) & $100 \times 100$ & \textbf{4.2 G} \\
\bottomrule
\end{tabular}
\caption{\textbf{Inference memory.} Inference GPU memory consumption was benchmark-tested with batch size of one on each GPU}
\label{tab: memory}
\end{table}

\textbf{Memory Efficiency.}
Because of the residual prediction strategy, we are able to adopt a lightweight occupancy head that enables efficient training and exhibits lower GPU memory usage, making our model training-friendly. Table~\ref{tab: memory} shows that InstanceBEV, which uses dense BEV and fully leverages the highly compressed instance-level information, consumes significantly less GPU memory during inference compared to other baselines relying on dense and sparse representations. This reduces the deployment burden in edge scenarios and making InstanceBEV inferencing-friendly. Moreover, since our model naturally generates instance-level outputs, it provides a highly compact scene representation that is well-suited for extension to other downstream tasks.

\subsection{Risk of Visible mask.} 
Camera mask and lidar visible mask are commonly used during evaluation to exclude voxels outside the visible field of view. However, these masks are generated from ground truth occupancy and are inherently unavailable during real-world deployment. This mismatch introduces a critical reliability gap between evaluation and real-world deployment. As shown in the visualization in Figure~\ref{fig: risk}, FB-Occ exhibits significant false positive predictions in regions that would have been masked out during evaluation—especially on road surfaces—leading to large, unrealistic occupied areas and posing serious safety risks in practical applications. In addition, camera masks encourage the model to overfit visible surfaces and produce thicker surfaces, which significantly hack mIoU scores that uses camera mask~\cite{liu_fully_2024}. Comparing to FB-Occ and etc., SparseOcc and InstanceBEV are trained without camera mask. This avoid encourage the model to overfit visible surfaces and produce thicker surfaces, which significantly hack mIoU scores that uses camera mask.

\section{Visualization}
\begin{figure*}[htbp]
\centering
\includegraphics[width=1.0\textwidth]{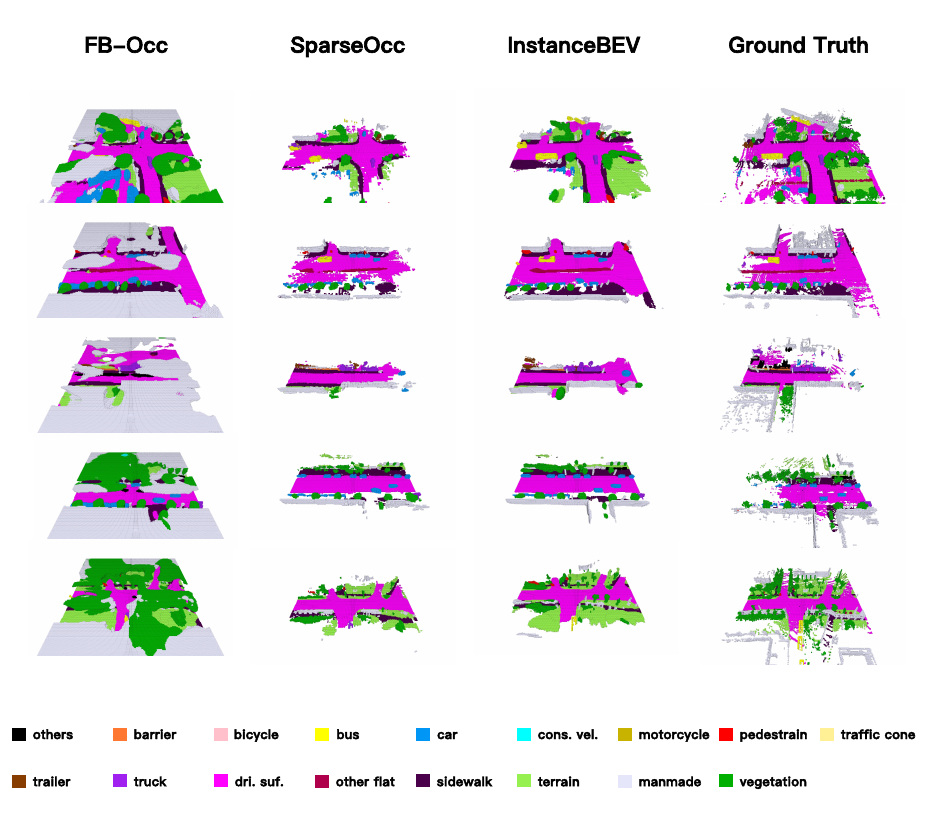}
\caption{\textbf{Occupancy visualization on 3D view.} FB-Occ produces occupancy outputs with substantial noise, while SparseOcc suffers from significant missed detections. In contrast, InstanceBEV maintains clean outputs while recovering rich details.}
\label{fig:occ vis}
\end{figure*}
\subsection{Occupancy Prediction Visualization}  
Figure~\ref{fig:occ vis} presents qualitative comparisons of 3D occupancy predictions from different models. Notably, FB-Occ does not utilize the camera mask during inference. As previously discussed, this mask is derived from ground-truth occupancy and therefore cannot be obtained during real-world deployment. As a result, FB-Occ tends to produce overestimated predictions, especially in regions not visible to any camera. This is reflected in the visualizations, where FB-Occ yields the noisiest outputs with numerous false positives in occluded or distant areas. In contrast, SparseOcc produces significantly cleaner predictions. However, its inherent sparsification leads to limited expressiveness in complex scenes, often manifesting as missing voxels. InstanceBEV strikes a better balance, producing clean and accurate predictions even in challenging scenarios.

\begin{figure*}[htbp]
\centering
\includegraphics[width=1.0\textwidth]{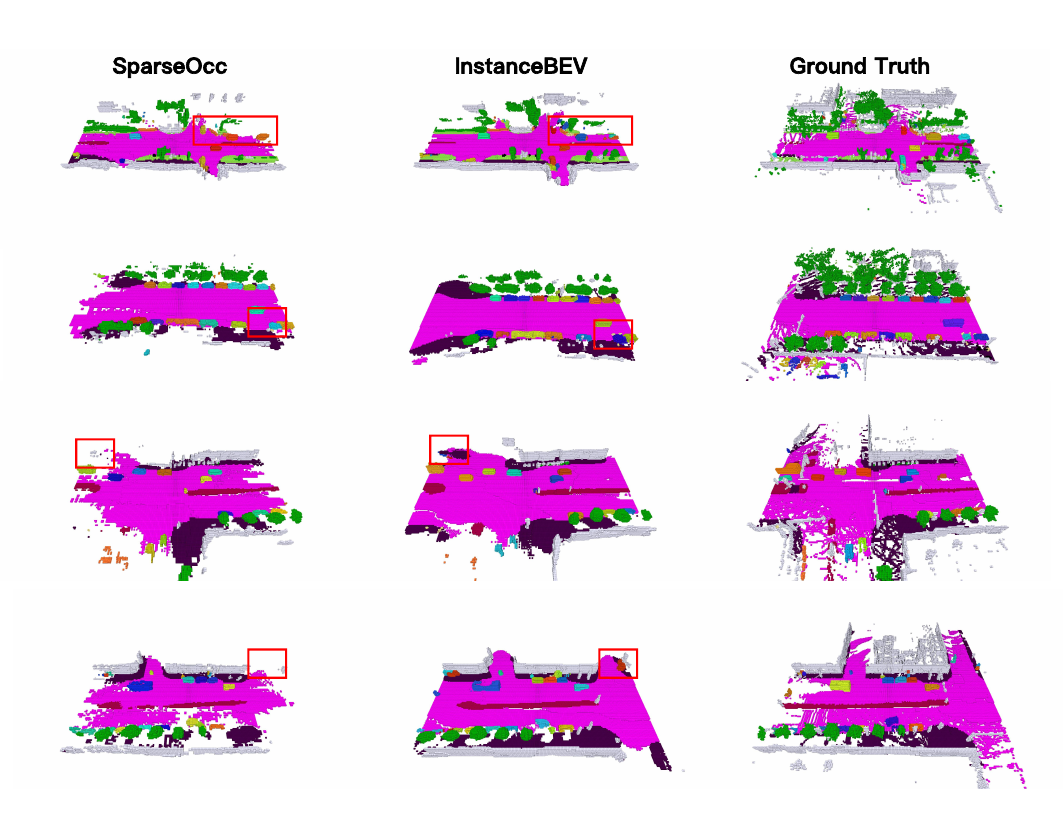}
\caption{\textbf{3D Panoptic Segmentation. Different instances are distinguished by colors.} SparseOcc shows limited representational capacity in complex scenes, often leading to discontinuous occupancy for large objects and missed detections for small objects.}
\label{fig:pano vis}
\end{figure*}

\subsection{Panoptic Segmentation Visualization}  
We visualize panoptic occupancy segmentation in Figure~\ref{fig:pano vis} by associating the occupancy predictions from InstanceBEV with the instance-level BEV masks. Benefiting from its dense BEV representation, InstanceBEV produces more complete and coherent predictions for large, continuous objects such as roads, as well as for structured objects like vehicles. In contrast, the sparsified representation used in SparseOcc may lead to deformations or incomplete predictions due to resolution. Moreover, the spatial sparsity in SparseOcc can introduce cumulative errors during scene completion, making it prone to missing some small objects.

\section{Conclusion}
In conclusion, we propose InstanceBEV to tackle the challenge of global modeling in BEV, which leverages a bidirectional cross-attention mechanism to jointly model BEV and instance features through once attention computation, while maintaining real-time performance. Additionally, we introduce the Residual Prediction strategy, enabling residual connections achieves dense connections performance without any cost.

InstanceBEV achieves significant performance improvements while ensuring low memory usage and real-time efficiency. Furthermore, the multi-space feature modeling of InstanceBEV demonstrates strong generalization potential and provides valuable insights for feature-sharing strategies and multi-task learning.


\bibliography{main}


 





\end{document}